\documentclass[letterpaper, 10 pt, conference]{ieeeconf}

\IEEEoverridecommandlockouts
\usepackage{placeins}
\let\labelindent\relax
\usepackage{enumitem}
\usepackage{color}
\usepackage[dvipsnames]{xcolor}
\usepackage{caption}
\usepackage{multirow}
\usepackage{csvsimple}
\usepackage{graphicx} 
\title{\LARGE \bf
Learning local trajectories for high precision robotic tasks : application to KUKA LBR iiwa Cartesian positioning}

\author{ Joris Gu\'erin, Olivier Gibaru, Eric Nyiri and St\'ephane Thiery  
\thanks{J. Gu\'erin, O. Gibaru, E. Nyiri and S. Thiery are with Laboratoire des Sciences de l'Information et des Syt\`emes (CNRS UMR 7296) INSM team, 8, Bd Louis XIV, 59046 Lille Cedex, France {\tt\small \{joris.guerin, olivier.gibaru, stephane.thiery,  eric.nyiri\}@ensam.eu}.}%
}

\begin{document}

\maketitle
\thispagestyle{empty}
\pagestyle{empty}

\begin{abstract}

To ease the development of robot learning in industry, two conditions need to be fulfilled. Manipulators must be able to learn high accuracy and precision tasks while being safe for workers in the factory. In this paper, we extend previously submitted work \cite{CDC16} which consist in rapid learning of local high accuracy behaviors. By exploration and regression, linear and quadratic models are learnt for respectively the dynamics and cost function. Iterative Linear Quadratic Gaussian Regulator combined with cost quadratic regression can converge rapidly in the final stages towards high accuracy behavior as the cost function is modelled quite precisely. In this paper, both a different cost function and a second order improvement method are implemented within this framework. We also propose an analysis of the algorithm parameters through simulation for a positioning task. Finally, an experimental validation on a KUKA LBR iiwa robot is carried out. This collaborative robot manipulator can be easily programmed into safety mode, which makes it qualified for the second industry constraint stated above. 
\end{abstract}

\section{INTRODUCTION}\label{introduction}

\subsection{Reinforcement Learning for Industrial Robotics}

In a close future, it is likely to see robots programming themselves to carry out new industrial tasks. From manufacturing to assembling, there are a wide range of tasks that can be performed faster and with higher accuracy by robot manipulators. Over the past two decades, reinforcement learning in robotics \cite{survey_RL_robotics, survey_PS_robotics} has made rapid progress and enabled robots to learn a wide variety of tasks \cite{interestRobotRL1, interestRobotRL2, interestRobotRL3}. The emergence of self-programming robots might speed up the development of industrial robotic platforms insofar as robots can learn to execute tasks with very high accuracy and precision.

One major step in a robot learning algorithm is the exploration phase. In such phase, random commands are sent to the robot such that it discovers both its environment and the way it responds to commands sent. In this process, random commands are sent to the robot, which can result in any possible movement within its reachable workspace. In an industrial context, such unpredictable behavior is dangerous, for instance when a robot has to learn a task jointly with a human worker (e.g. an assembly task). For this reason, it seems interesting to work with KUKA LBR iiwa robot manipulators, which are very good for collaborative tasks as their compliance can be adjusted easily and they can be programmed to stop when feeling contact.
 
\subsection{Literature Overview}

Reinforcement Learning (RL) and Optimal Feedback Control (OFC) are very similar in their formulation : find a policy that minimizes a certain cost function under a certain dynamics (see section \ref{section2} for more details). They both enable to phrase many challenging robotic tasks. Indeed, a solution to such problem is both an optimal open-loop trajectory and a feedback controller.  If the dynamics is linear and the cost function quadratic, an optimal solution can be computed analytically using Linear-Quadratic-Regulators theory \cite{optimal_control_book}. 

When the dynamics is more complex (non-linear), the problem becomes more difficult but can still be solved with iterative Linear-Quadratic-Regulator algorithm (iLQR or iLQG) \cite{iLQG2}. As its name suggests, this algorithm iteratively fits local linear approximations to the dynamics and computes a locally optimal solution under this linear model. In the context of RL, the dynamics is considered unknown. To deal with this issue, \cite{mitrovic, GPS_unknown_dynamics_simulation} have proposed to build the linear model by exploring the environment and make a linear regression.

In \cite{CDC16}, we recently proposed another method that consists in computing the cost function likewise, using exploration and quadratic regression. This way, the model is more precise and can converge faster towards high precision tasks, which is the main purpose of our research. Indeed, in some tasks, for example Cartesian positioning, a typical approach \cite{GPS_unknown_dynamics_robot} consists in including the Cartesian position in the state, build a linear model and then build a quadratic cost from this linear approximation. Such approach does not really make sens as this quantity has already been approximated in the first order and thus cannot produce a good second order model for update.

\subsection{Main contribution and paper organization}

In this paper, we extend the concepts of \cite{CDC16}. Second order methods have been implemented to compute trajectory update and this way increase the speed of convergence by reducing the number of iLQG pass required. We also study the influence of different parameters on the speed of convergence.Such parameters are compared and chosen using the V-REP software \cite{VREP}. Finally, we propose an experimental validation on the physical device, using the parameters found by simulation. The KUKA LBR iiwa learns a positioning task in Cartesian space using angular position control without any geometric model provided. This rather simple task enables to measure easily the accuracy of the policy found.

The paper is organized as follows. Derivation of iLQG  with learnt dynamics and cost function is written in section \ref{section2}. In section \ref{section3}, we try to find the best learning parameters through simulating the same learning situation with different parameters using the VREP simulation software. Experimental validation on KUKA LBR iiwa is presented in section \ref{section4} and section \ref{discussion} proposes a discussion on the results and future work.

\section{LEARNING A LOCAL TRAJECTORY WITH HIGH PRECISION}\label{section2}

This section summarizes the method used. First, the derivation of iLQG in the context of unknown dynamics and learnt cost function is written. The second order method to compute the improved controller is explain in a second step.

\subsection{A few definitions}\label{subsection21}

\begin{figure}[t]
    \centering
    \includegraphics[width=3.5cm, height=4.62cm]{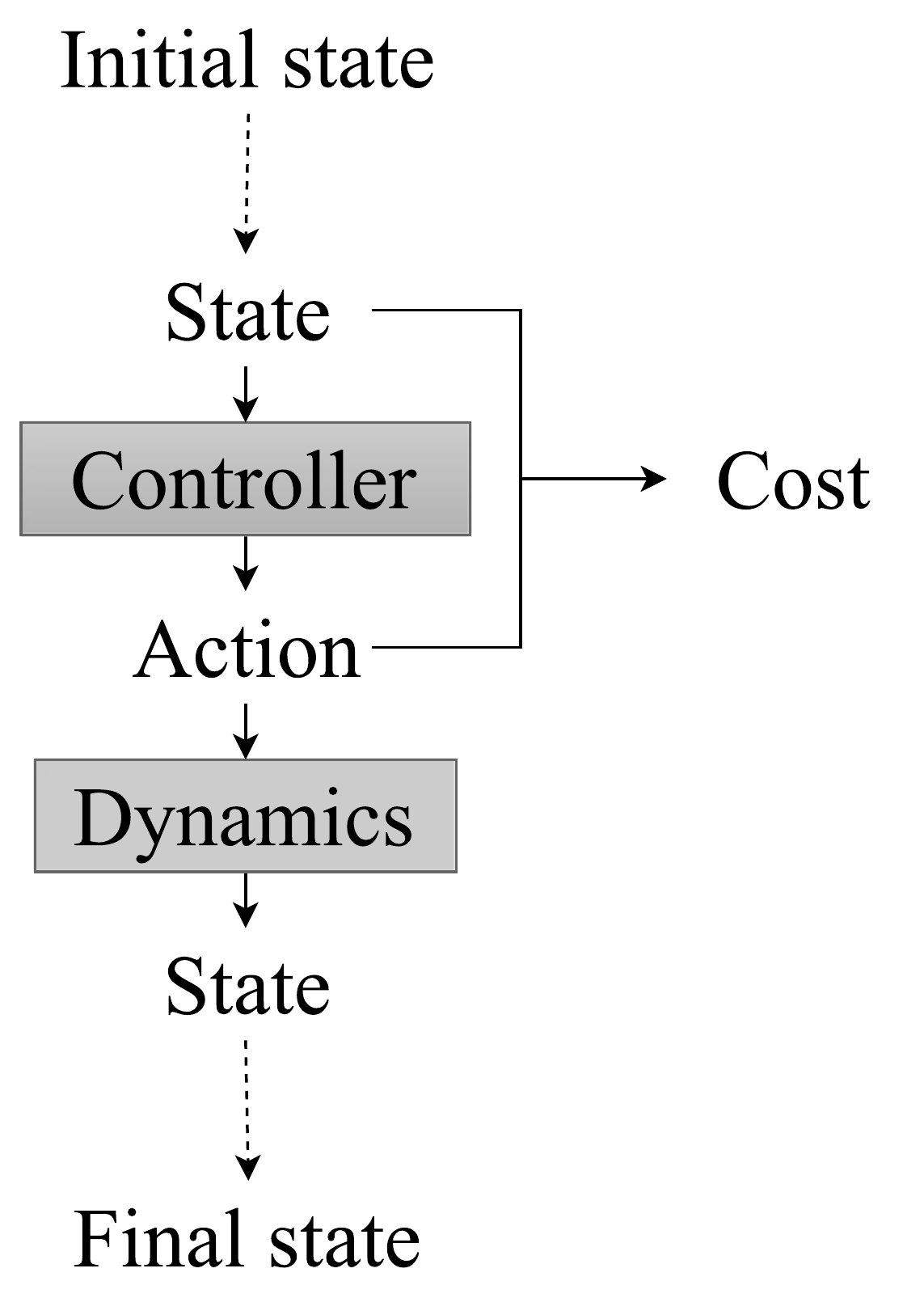}
    \caption{Definition of a trajectory}
    \label{trajectory_definition}
\end{figure}

This section begins with some useful definitions:
\begin{itemize}
    \item A \textbf{trajectory} $\tau$ of length $T$ is defined by the repetition $T$ times of the pattern shown in Fig. \ref{trajectory_definition}. Mathematically, it can be denoted by
    \begin{displaymath}
        \{\mathbf{x}_0, \mathbf{u}_0,  \mathbf{x}_1, \mathbf{u}_1, ..., \mathbf{u}_{T-1}, \mathbf{x}_{T}\},
    \end{displaymath}
    where $\mathbf{x}_t$ and $\mathbf{u}_t$ represent respectively state and control vectors. In our problem (see section \ref{section3} and \ref{section4}), the state is the vector of joint positions and the actions are joints target positions.
    \vspace{5pt}
    \item The \textbf{cost} and \textbf{dynamics} functions are defined as follows:
    \begin{equation}
        \label{general_cost}
        l_t = L_t(\mathbf{x_t}, \mathbf{u}_t),
    \end{equation}
    \begin{equation}
        \label{general_dynamics}
        \mathbf{x}_{t+1} = F_t(\mathbf{x}_t, \mathbf{u}_t).
    \end{equation}
    $L_t$ outputs the cost and $F_t$ the next state, both with respect to previous state and action.
    \vspace{5pt}
    \item The \textbf{controller} is the function we want to optimize. For a given state, it needs to output the action with smallest cost that follows the dynamics. In our case, it is denoted by $\Pi$ and has the special special form of a time-varying linear controller:
    \begin{equation}
    \label{global_controller}
        \mathbf{u}_t = \Pi(\mathbf{x}_t) = K_t \mathbf{x}_t + k_t.
    \end{equation}
\end{itemize}

The guiding principle of iLQG is to alternate between the two following steps. From a nominal trajectory, denoted by $\mathbf{\bar{x}}_t$ and $\mathbf{\bar{u}}_t$, $ t \in \{0, ..., T\}$,  compute a new local optimal controller. From a given controller, draw a new nominal trajectory.

\subsection{Local approximations of cost and dynamics}\label{subsection22}

As explained earlier, from a given nominal trajectory $\bar{\tau}$, the goal is to update the controller such that the cost is minimized in its neighborhood. In this process, the first step is to compute local approximations of the cost function and dynamics around the nominal trajectory:
\begin{equation}
\label{local_dynamics}
    F_t(\mathbf{\bar{x}}_t + \mathbf{\delta x}_t, \mathbf{\bar{u}}_t + \mathbf{\delta u}_t) =  \mathbf{\bar{x}}_{t+1} + F_{xu_t} \mathbf{\delta xu}_t,
\end{equation}
\begin{IEEEeqnarray}{l}
\label{local_cost}
    L_t(\mathbf{\bar{x}}_t + \mathbf{\delta x}_t, \mathbf{\bar{u}}_t + \mathbf{\delta u}_t) = \bar{l}_t + L_{xu_t} \mathbf{\delta xu}_t + \nonumber\\
    \frac{1}{2} \mathbf{\delta xu}_t^T L_{xu,xu_t} \mathbf{\delta xu}_t,
\end{IEEEeqnarray}
where $\mathbf{\delta x}_t$ and $\mathbf{\delta u}_t$ represent variations from the nominal trajectory and $xu_t$ is the vector $[x_t,u_t]^T$. The notation $A_z$ (resp. $A_{z_1,z_2}$) is the Jacobian (resp. Hessian) matrix of $A$ w.r.t. $z$ (resp. $z_1$ and $z_2$).

We propose to compute both approximations following an exploration and regression scheme. The first stage generates a certain number $N$ of random trajectories around the nominal. These trajectories are normally distributed around $\bar{\tau}$ with a certain time-varying covariance $\Sigma_t$. Hence, during the sample generation phase, the controller is stochastic and follows:
\begin{equation}
\label{global_controller}
    P(\mathbf{u}_t|\mathbf{x}_t) = \mathcal{N}(K_t \mathbf{x}_t + k_t, \Sigma_t), \hspace{5pt} \forall t \in \{0, ..., T\},
\end{equation}
where $\mathcal{N}$ stands for normal distribution. From these samples, we can make two regressions a linear one to get the dynamics and a second order polynomial one \cite{polyreg} to approximate the cost function.

\subsection{Update the controller}

This section is about updating the controller once we have good Taylor expansions of the dynamics and cost function. In order to get a controller with low cost over the whole trajectory, we need to use the two value functions: $Q$ and $V$. $Q^{\Pi}_t$ represents the expected cost until the end of the trajectory if following $\Pi$ after being in state $x_t$ and selecting action $u_t$. $V^{\Pi}_t$ is the same but conditioned only on $x_t$, if $\Pi$ is deterministic, these two functions are exactly the same. The reader can refer to \cite{RL_textbook} for more detailed definitions of these value functions.

First, we need to compute quadratic Taylor expansions of both value functions:
\begin{IEEEeqnarray}{l}
\label{local_Q}
    Q^{\Pi}_t(\mathbf{\bar{x}}_t + \mathbf{\delta x}_t, \mathbf{\bar{u}}_t + \mathbf{\delta u}_t) =  Q_{0_t} + Q_{xu_t} \mathbf{\delta xu}_t + \nonumber\\
    \frac{1}{2} \mathbf{\delta xu}_t^T Q_{xu,xu_t} \mathbf{\delta xu}_t,
\end{IEEEeqnarray}
\begin{equation}
\label{local_V}
    V^{\Pi}_t(\mathbf{\bar{x}}_t + \mathbf{\delta x}_t) =  V_{0_t} + V_{xu_t} \mathbf{\delta xu}_t + \frac{1}{2} \mathbf{\delta xu}_t^T V_{xu,xu_t} \mathbf{\delta xu}_t.
\end{equation}

In the context of trajectory optimization defined above, \cite{GPS_unknown_dynamics_robot} shows that these two functions can be approximated quadratically by

\begin{IEEEeqnarray}{l}
\label{new_Q}
    Q_{xu,xu_t} = L_{xu,xu_t} + F_{xu_t} V_{x,x_{t+1}} F_{xu_t}^T, \nonumber\\
    Q_{xu_t} = L_{xu_t} + V_{x_{t+1}} F_{xu_t}^T, \nonumber\\
    V_{x,x_t} = Q_{x,x_t} + Q_{u,x_t}^T Q_{u,u_t}^{-1} Q_{u,x_t}\nonumber\\
    V_{x_t} = Q_{x_t} + Q_{u,x_t}^T Q_{u,u_t}^{-1} Q_{u_t}^T.
\end{IEEEeqnarray}
These functions are computed backward for all the time steps, starting with $V_T = l_T(x_T)$, the final cost.

Under such quadratic value functions, following the derivation in \cite{iLQG}, we can show that the optimal controller under such dynamics and cost is defined by
\begin{IEEEeqnarray}{l}
\label{new_Q}
    K_t=-Q_{u,u_t}^{-1} Q_{u,x_t}, \nonumber\\
    k_t=\bar{u_t} -Q_{u,u_t}^{-1} Q_{u_t} - K_t \bar{x_t}.
\end{IEEEeqnarray}

A criterion to compute the new covariance is also needed. The goal being to explore the environment, we follow \cite{GPS} and choose the covariance with highest entropy in order to maximize information gained during exploration. Such covariance matrix is:
\begin{equation}
    \Sigma_t = Q_{u,u_t}^{-1}.
\end{equation}

\subsection{Limit the deviation from nominal trajectory}

The controller derived above is optimal only if the dynamics and cost are respectively linear and quadratic everywhere. The approximations being only valid locally, the controller needs to be kept close from the nominal trajectory to remain acceptable for update. This problem can be solved by adding a constraint to the cost minimization problem:
\begin{equation}
\label{KL_constraint}
    D_{KL}(p_{new}(\tau)||p_{old}(\tau)) \leq \epsilon,
\end{equation}
where $D_{KL}$ is the statistical Kullback-Leibler divergence. $p_{old}(\tau)$ and $p_{new}(\tau)$ are the probability trajectory distributions under the current controller and the updated one.

\cite{GPS_unknown_dynamics_robot} shows that such constrained optimization problem can be solved rather easily by introducing the modified cost function: 
\begin{equation}
    l_{mod}(x_t, u_t) = \frac{1}{\eta} l(x_t, u_t) - \log(p_{old}(x_t, u_t))
\end{equation}
Indeed, using dual gradient descent, we can find a solution to the constrained problem by alternating between the two following steps: \begin{itemize}
    \item Compute the optimal unconstrained controller under $l_{mod}$ for a given $\eta$
    
    \item If the controller does not satisfy (\ref{KL_constraint}), increase $\eta$.
\end{itemize} 
A large $\eta$ has the effect of increasing the importance on constraint satisfaction, so the larger $\eta$ is, the closer the new trajectory distribution will be from the previous one.

\subsection{Initialize $\eta$ and choose $\epsilon$}

The way $Q$ is defined from approximation does not guaranty positive definiteness for $Q_{u,u_t}$. Which means that it might not be eligible to be a covariance matrix. This issue is addressed by increasing $\eta$ such that the distribution is close enough from the previous one. As the previous trajectory has a positive definite covariance, there must be an $\eta$ that will enforce positive definiteness. This gives a good way to initialize $\eta$ for a given pass.

Finally, the choice of $\epsilon$ is very important. If it is too small, the controller sequence won't progress towards optimality. On the other hand, if it is too large, it might be unstable. The idea is to start with a certain $\epsilon_{ini}$ and decrease it if the new accepted controller is worst than the previous one.

\section{KUKA LBR IIWA POSITIONING: TUNE THE LEARNING PARAMETERS}\label{section3}

A validation of the method is proposed on learning a simple inverse kinematics task. We consider a KUKA LBR iiwa robot (Fig. \ref{iiwa_VREP}), where the geometric parameters are unknown. The state variables are the joints angular positions and the control vector gathers target joints positions for next state. The idea is to reach a Cartesian position of the end effector with high accuracy ($<0.1mm$) without any geometric model.

\subsection{Cost function}

For this problem, the cost function needs to be expressed in terms of the Cartesian distance between the end-effector and the target point. We chose the cost function proposed in \cite{GPS_unknown_dynamics_robot}:
\begin{equation}
    l(d)=d^2 + v \log(d^2 + \alpha),
\end{equation}
where $v$ and $\alpha$ are both real user defined parameters. As we do not consider any geometric parameter of the robot, the distance cannot be obtained with direct model considerations and needs to be measured from sensors.

\subsection{Tune the algorithm parameters}

\begin{figure}[t]
    \centering
    \includegraphics[width=4cm, height = 5.6cm]{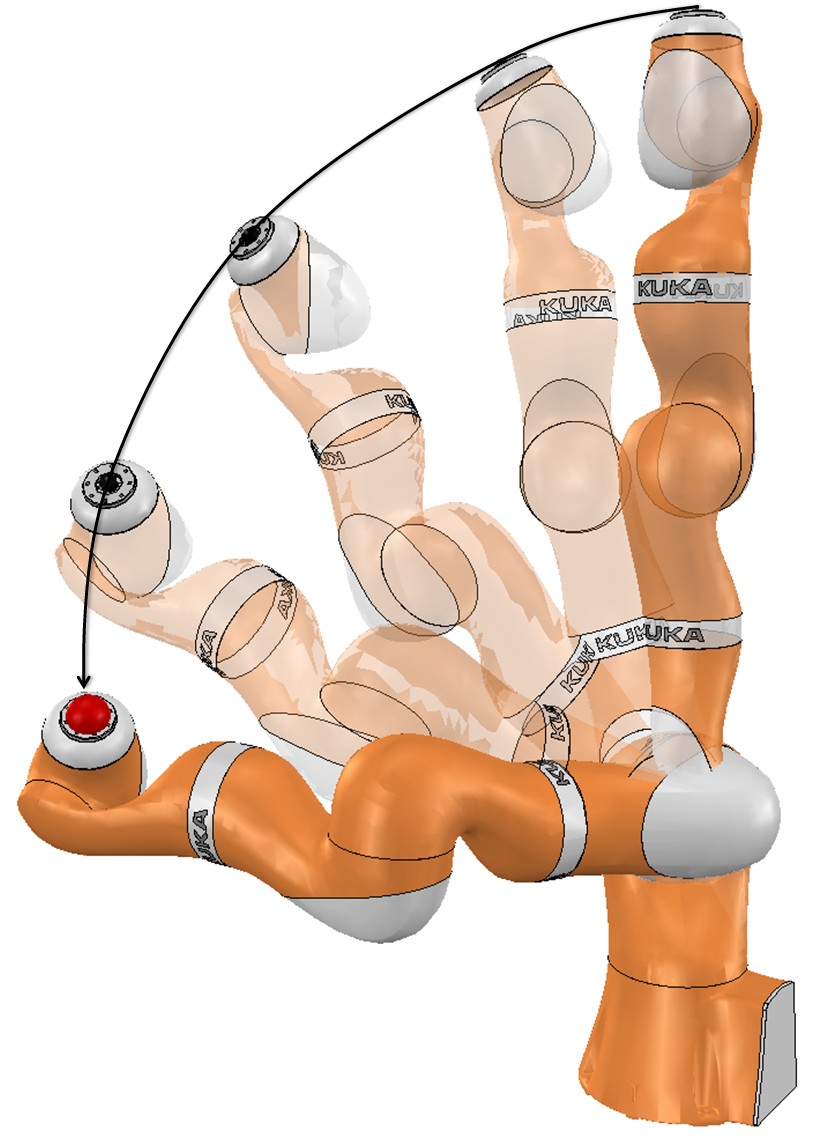}
    \caption{Trajectory learnt on V-REP software with a KUKA LBR iiwa}
    \label{iiwa_VREP}
\end{figure}

Previous work \cite{CDC16} showed that a number of samples around $40$ is a good balance between accurate quadratic regression and exploration time for $7$ d.o.f. robots. So we carry out our experiments with $N=40$. Among all the parameters defined in previous sections, we identified $4$ critical ones : $cov_{ini}$ (the initial covariance, defined below), $v$ and $\alpha$ from the cost function, $\epsilon_{ini}$. In this section, we learned optimal angular positions for the situation below with different sets of values on these parameters using V-REP simulation software. The situation is the following:
\begin{itemize}
   \item Initial position : All $7$ angles at $0$ (straight position on Fig. \ref{iiwa_VREP})

    \item Target position : Cartesian vector $[500, 500, 500]^T$ in $mm$, in the robot frame (red sphere on Fig. \ref{iiwa_VREP})
   
    \item Initial mean command : target angular positions = initial positions (no move command).
\end{itemize}
Fig. \ref{iiwa_VREP} show a trajectory found by the algorithm.

The initial covariance matrix is also an important parameter as it defines the exploration range for all the future steps. Indeed, if it has large values, next iteration needs to have large covariance also because of (\ref{KL_constraint}). In our implementation, we start with diagonal covariance matrix where all the diagonal entries are the same. we denote $cov_{ini}$ the initial value of such diagonal entries, it is one of the parameters to be studied.

\subsection{Results and analysis}

From what we acknowledged running our algorithm, we picked up three values for each parameter and tried all the $81$ possible combinations to choose a good set of parameters for positioning task. Results obtained are summarized in table \ref{table_result}. In our simulation, the robot was only allowed $16$ trials to reach $0.1mm$ precision. Thus, we insist that in table \ref{table_result}, an underlined number represents the number of iLQG iterations before convergence whereas other numbers are the remaining distance to objective after $16$ iterations.

\begin{table}[ht]
\caption{Algorithm parameters influence}
\begin{center}

\begin{tabular}{c|c|c|c|c}
\multicolumn{5}{c}{$cov_{ini}=1$}\tabularnewline
\hline
\multirow{2}{*}{$v$} & \multirow{2}{*}{$\epsilon_{ini}$} & \multicolumn{3}{c}{$\alpha$}\tabularnewline
 & & $10^{-3}$ & $10^{-5}$ & $10^{-7}$\tabularnewline
\hline
\multirow{3}{*}{$0.1$} & \textcolor{Black}{100} & \underline{\textcolor{Black}{11}} & \underline{\textcolor{Black}{16}} & \underline{\textcolor{Black}{13}}\tabularnewline
 & \textcolor{gray}{1000} & \textcolor{gray}{0.25} & \underline{\textcolor{gray}{12}} & \underline{\textcolor{gray}{10}}\tabularnewline
 & \textcolor{Blue}{10000} & \underline{\textcolor{Blue}{13}} & \textcolor{Blue}{0.27} & \underline{\textcolor{Blue}{8}}\tabularnewline
\cline{1-5}
 \multirow{3}{*}{$1$} & \textcolor{Black}{100} & \textcolor{Black}{0.11} & \underline{\textcolor{Black}{14}} & \underline{\textcolor{Black}{16}}\tabularnewline
 & \textcolor{gray}{1000} & \underline{\textcolor{gray}{10}} & \underline{\textcolor{gray}{12}} & \underline{\textcolor{gray}{10}}\tabularnewline
 & \textcolor{Blue}{10000} & \textcolor{Blue}{0.10} & \textcolor{Blue}{1.69} & \textcolor{Blue}{0.24}\tabularnewline
\cline{1-5}
 \multirow{3}{*}{$10$} & \textcolor{Black}{100} & \textcolor{Black}{0.11} & \textcolor{Black}{0.22} & \textcolor{Black}{0.84}\tabularnewline
 & \textcolor{gray}{1000} & \textcolor{gray}{0.13} & \underline{\textcolor{gray}{12}} & \textcolor{gray}{0.20}\tabularnewline
 & \textcolor{Blue}{10000} & \underline{\textcolor{Blue}{13}} & \textcolor{Blue}{0.23} & \underline{\textcolor{Blue}{15}}\tabularnewline
\hline
\end{tabular}

\vspace*{2.5mm}

\begin{tabular}{c|c|c|c|c}
\multicolumn{5}{c}{$cov_{ini}=10$}\tabularnewline
\hline
\multirow{2}{*}{$v$} & \multirow{2}{*}{$\epsilon_{ini}$} & \multicolumn{3}{c}{$\alpha$}\tabularnewline
 & & $10^{-3}$ & $10^{-5}$ & $10^{-7}$\tabularnewline
\hline
\multirow{3}{*}{$0.1$} & \textcolor{Black}{100} & \textcolor{Black}{0.32} & \textcolor{Black}{0.15} & \textcolor{Black}{0.39}\tabularnewline
 & \textcolor{gray}{1000} & \textcolor{gray}{0.45} & \textcolor{gray}{0.28} & \textcolor{gray}{0.22}\tabularnewline
 & \textcolor{Blue}{10000} & \textcolor{Blue}{0.30} & \textcolor{Blue}{0.29} & \textcolor{Blue}{0.31}\tabularnewline
\cline{1-5}
 \multirow{3}{*}{$1$} & \textcolor{Black}{100} & \textcolor{Black}{0.14} & \textcolor{Black}{0.32} & \textcolor{Black}{0.32}\tabularnewline
 & \textcolor{gray}{1000} & \underline{\textcolor{gray}{14}} & \textcolor{gray}{1.93} & \textcolor{gray}{1.70}\tabularnewline
 & \textcolor{Blue}{10000} & \textcolor{Blue}{1.82} & \textcolor{Blue}{0.99} & \textcolor{Blue}{0.11}\tabularnewline
\cline{1-5}
 \multirow{3}{*}{$10$} & \textcolor{Black}{100} &\textcolor{Black}{0.34} & \textcolor{Black}{0.38} & \textcolor{Black}{0.39}\tabularnewline
 & \textcolor{gray}{1000} & \textcolor{gray}{0.71} & \textcolor{gray}{0.29} & \textcolor{gray}{0.53}\tabularnewline
 & \textcolor{Blue}{10000} & \textcolor{Blue}{0.70} & \textcolor{Blue}{0.14} & \textcolor{Blue}{2.31}\tabularnewline
\hline
\end{tabular}

\vspace*{2.5mm}

\begin{tabular}{c|c|c|c|c}
\multicolumn{5}{c}{$cov_{ini}=100$}\tabularnewline
\hline
\multirow{2}{*}{$v$} & \multirow{2}{*}{$\epsilon_{ini}$} & \multicolumn{3}{c}{$\alpha$}\tabularnewline
 &  & $10^{-3}$ & $10^{-5}$ & $10^{-7}$\tabularnewline
\hline
\multirow{3}{*}{$0.1$} & \textcolor{Black}{100} & \textcolor{Black}{12.79} & \textcolor{Black}{12.42} & \textcolor{Black}{17.83}\tabularnewline
 & \textcolor{gray}{1000} & \textcolor{gray}{4.42} & \textcolor{gray}{0.30} & \textcolor{gray}{3.50}\tabularnewline
 & \textcolor{Blue}{10000} & \textcolor{Blue}{2.88} & \textcolor{Blue}{10.93} & \textcolor{Blue}{2.60}\tabularnewline
\cline{1-5}
 \multirow{3}{*}{$1$} & \textcolor{Black}{100} & \textcolor{Black}{24.37} & \textcolor{Black}{15.75} & \textcolor{Black}{10.13}\tabularnewline
 & \textcolor{gray}{1000} & \textcolor{gray}{7.66} & \textcolor{gray}{6.32} & \textcolor{gray}{1.87}\tabularnewline
 & \textcolor{Blue}{10000} & \textcolor{Blue}{2.67} & \textcolor{Blue}{8.37} & \textcolor{Blue}{6.44}\tabularnewline
\cline{1-5}
 \multirow{3}{*}{$10$} & \textcolor{Black}{100} & \textcolor{Black}{1.93} & \textcolor{Black}{8.93} & \textcolor{Black}{$10^{11}$}\tabularnewline
 & \textcolor{gray}{1000} & \color{gray}{8.03} & \textcolor{gray}{2.23} & \textcolor{gray}{3.50}\tabularnewline
 & \textcolor{Blue}{10000} & \textcolor{Blue}{2.70} & \textcolor{Blue}{4.83} & \textcolor{Blue}{2.60}\tabularnewline
\hline
\end{tabular}
\\[5pt]
\label{table_result}
\caption*{
\textit{
An underlined figure represents the number of iLQG iterations to reach $0.1mm$ precision, Other numbers represent the distance remaining after $16$ iterations.}}
\end{center}
\end{table}

\begin{figure}[tb]
    \centering
    \includegraphics[width=2.5in]{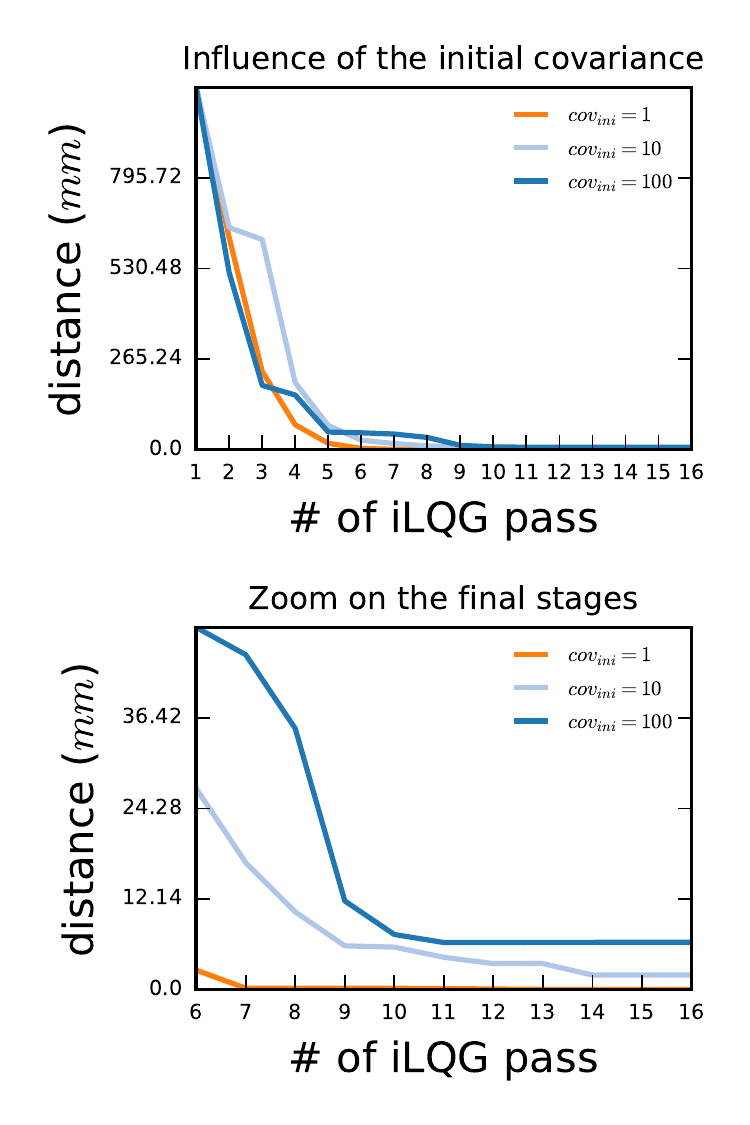}
    \caption{Covariance variation for $v=1$, $\alpha=10^{-5}$, $\epsilon_{ini}=1000$}
    \label{cov_curve}
\end{figure}

Together with the raw data in table \ref{table_result}, we plot the evolution of the distance within the iterations of a simulation for several sets of parameters. Looking at table \ref{table_result}, it seems that the most critical parameter is $cov_{ini}$. Fig. \ref{cov_curve} shows three learning curves where only $cov_{ini}$ varies. From here it appears that the initial covariance is not crucial in the early stages of the learning process. However, looking at the bottom plot, which is a zoom on the final steps, we acknowledge that if the covariance is too large, the algorithm will not converge towards the desired accuracy behavior. Hence, we recommend to keep $cov_{ini}$ around $1$ to obtain the desired accurate behavior.

\begin{figure}[tb]
    \centering
    \includegraphics[width=2.7in]{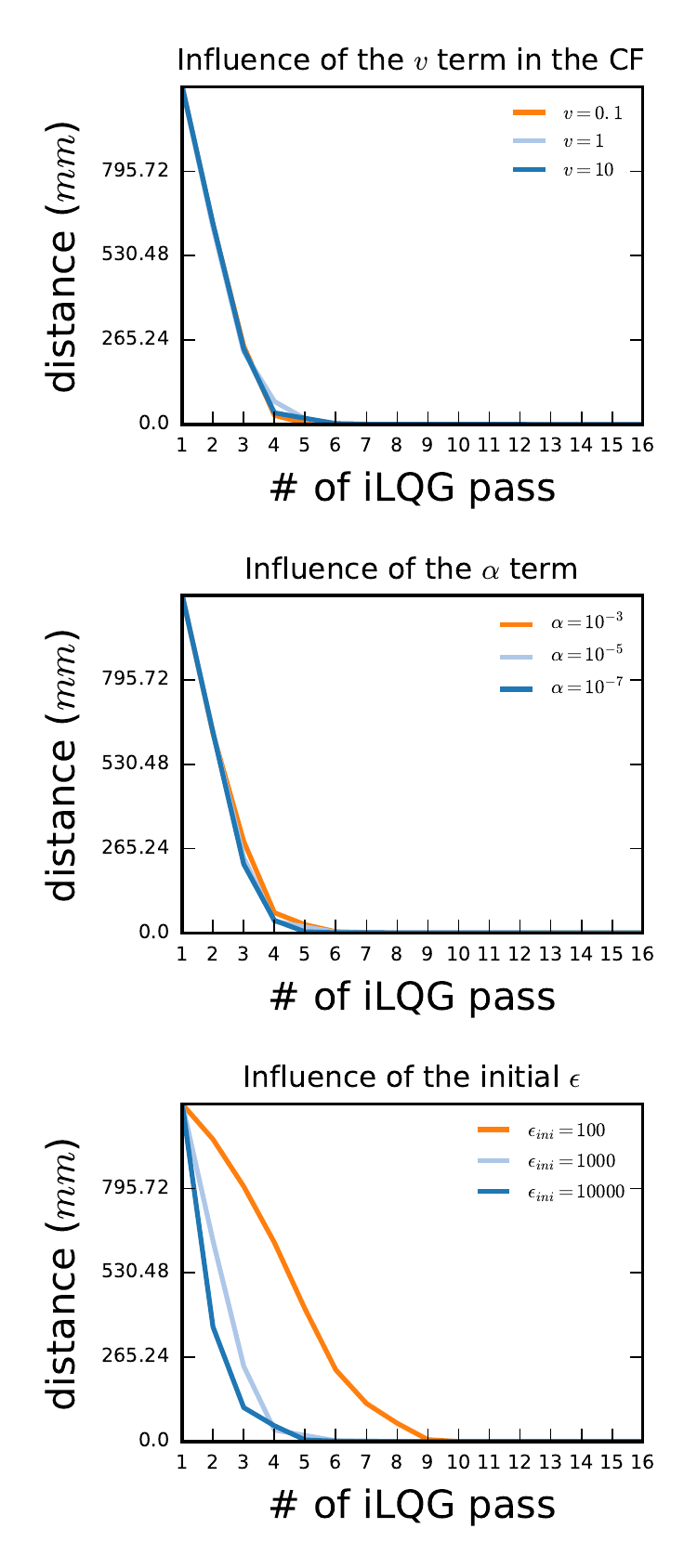}
    \caption{Variation of other parameters for $cov_{ini}=1$. When they do not vary, other parameters take the following values : $v=1$, $\alpha=10^{-5}$, $\epsilon_{ini}=1000$}
    \label{other_curves}
\end{figure}

After setting $cov_{ini}$ to $1$, we made the same plots for the other parameters (Fig. \ref{other_curves}). These reveal that $v$ and $\alpha$ do not appear to influence the behavior in this range of values. However, looking at the bottom plot, we can see that $\epsilon_{ini}$ needs to be kept large enough such that an iLQG iteration can make enough progress towards optimality. For small $\epsilon_{ini}$, we waste time stuck near the initial configuration. For the three plots in Fig. \ref{other_curves}, the zooms are not included as they do not reveal anything more.

Results in table \ref{table_result} seem to correspond to what has been said above. Hence, for the experimental validation in section \ref{section4}, we choose the configuration with smallest number of iterations: $cov_{ini}=1$, $v=0.1$, $\alpha=10^{-7}$ and $\epsilon_{ini}=10000$.

\section{EXPERIMENTAL VALIDATION ON THE ROBOT}\label{section4}

\begin{figure}[!tb]
    \centering
    \includegraphics[width=4cm, height = 4.8cm]{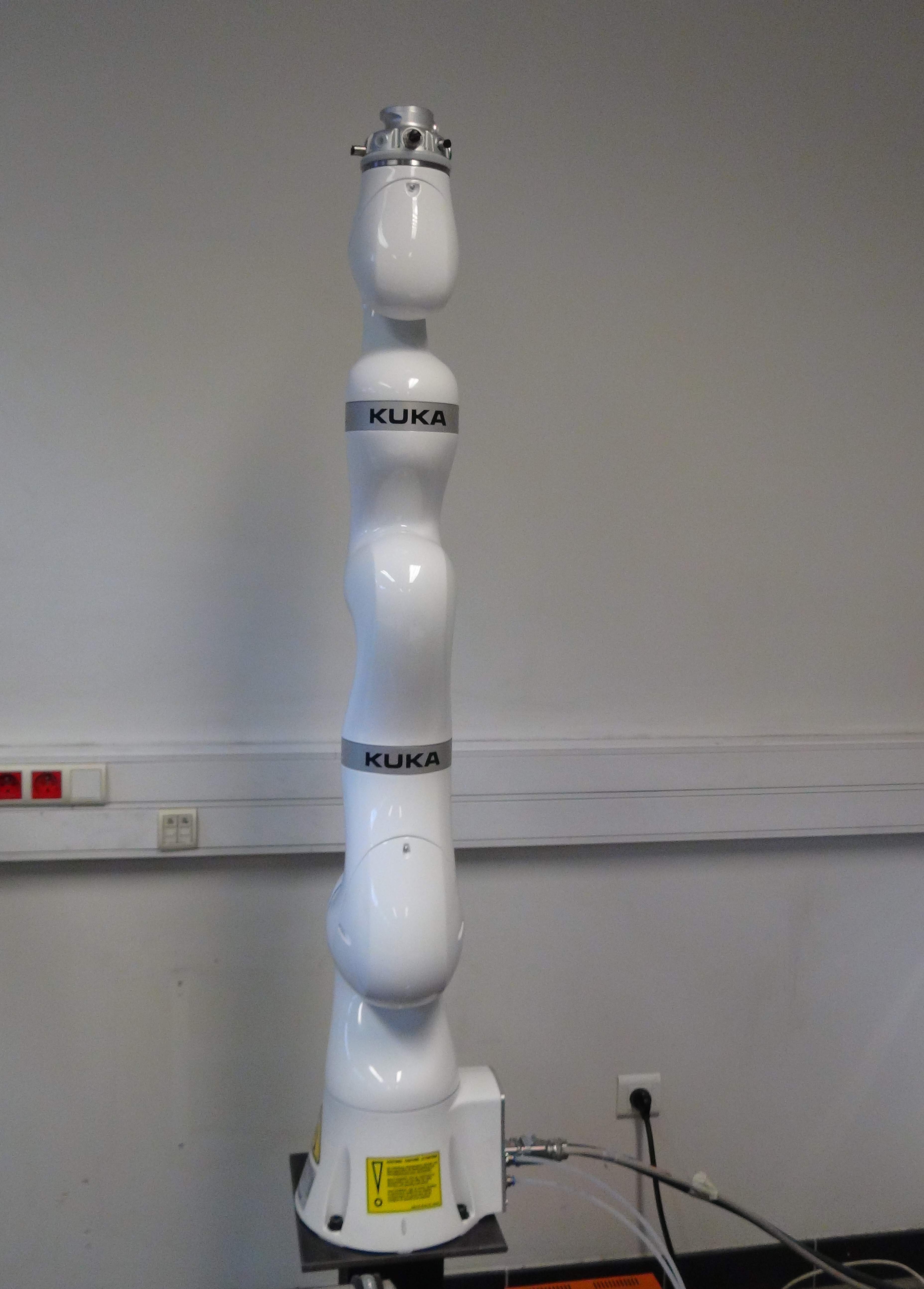}
    \includegraphics[width=4cm, height = 4.8cm]{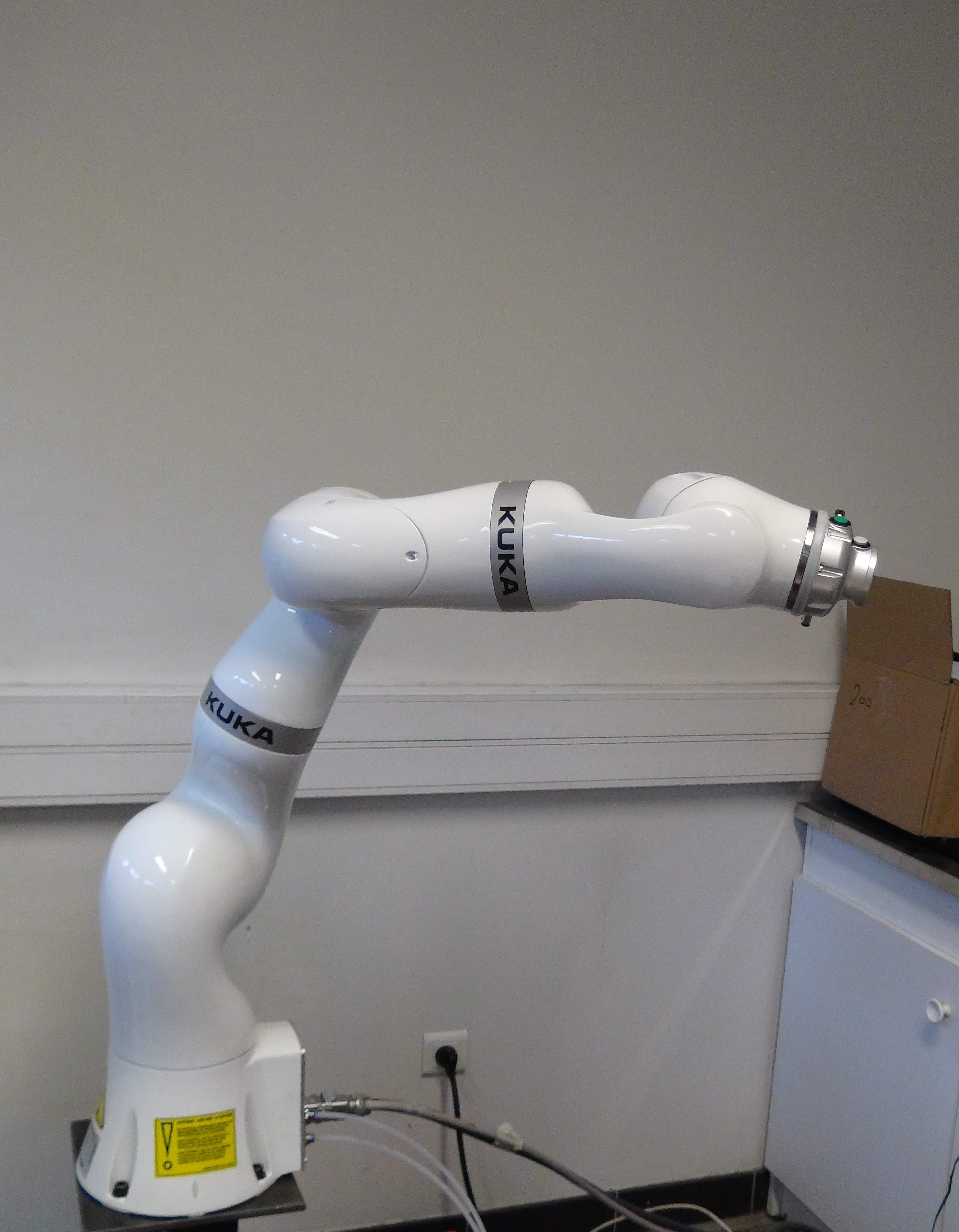}
    \caption{Initial configuration of the KUKA LBR iiwa.}
    \label{iiwa_pictures}
\end{figure}

In this section, we run our algorithm on a real KUKA LBR iiwa for a similar positioning task. The situation is slightly different: 
\begin{itemize}
    \item Initial position : $[140,0,0,0,0,0,0]^T$, angular positions in $^{\circ}$ (Fig. \ref{iiwa_pictures}, left picture)
    
    \item Target position : $[-600,400,750]^T$, Cartesian position, in $mm$ and in the robot frame.
    
    \item Initial mean command : target angular positions = initial positions (no move command).
\end{itemize}

The choice of changing the initial configuration was motivated by two reasons. First, it enables to show that the parameters found in section \ref{section3} are not case dependant. Second, we had constraints with our working environment and this configuration was better regarding research going on with other robots simultaneously.

Fig. \ref{iiwa_pictures} shows the KUKA LBR iiwa in its initial and final configuration (after reaching the desired end-effector position).

\subsection{Results obtained on the robot}

\begin{figure}[tb]
    \centering
    \includegraphics[]{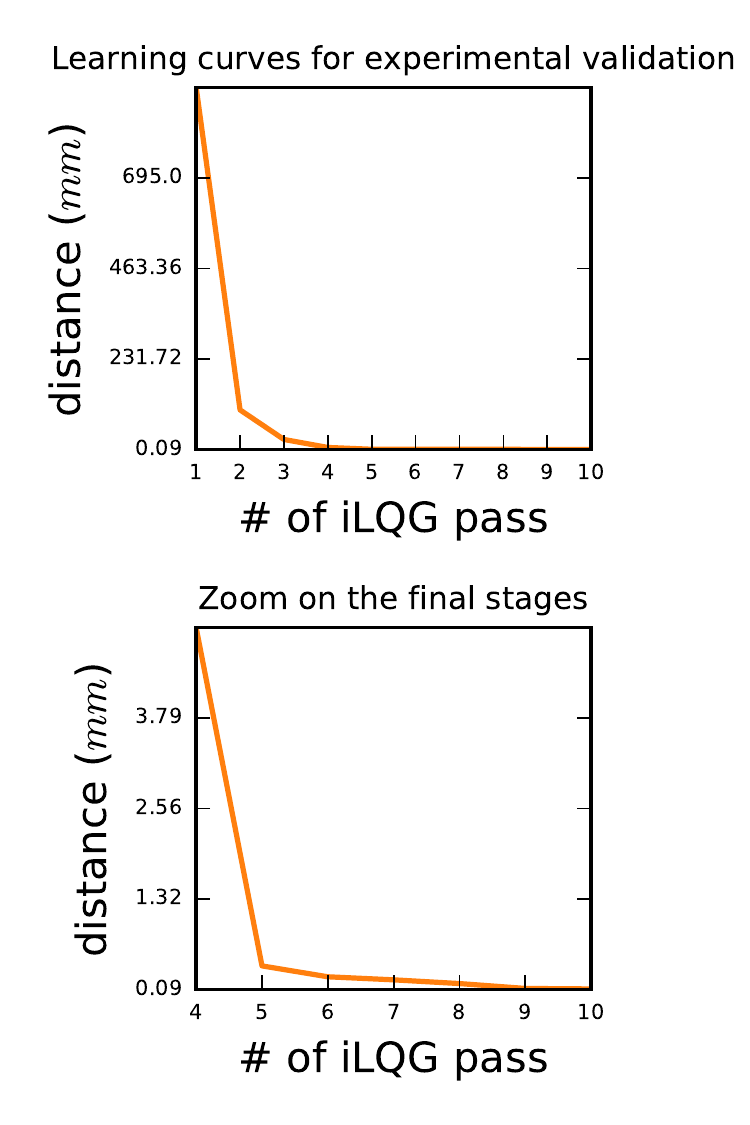}
    \caption{Learning curve iLQG on the robot.}
    \label{LCrobot}
\end{figure}

The learning process defined above resulted in the learning curve on Fig. \ref{LCrobot}. We note that it takes as many steps to go from initial configuration to $1mm$ accuracy than from $1mm$ to $0.1mm$. The final command provided by the algorithm is $[144.266,25.351,2.328,-56.812,5.385,24.984,4.754]^T$. Regarding the learning time, the overall process took approximately $9$ minutes, $6$ for exploration and $3$ for calculations.

\subsection{Measure the Cartesian distance}\label{measurements}

On this experimental validation, the distance was computed from the end-effector position read from the robot internal values. Even if it was probably computed thanks to direct DH model, our algorithm used it as an output from a black box. Thus, similar results would have been obtained using any other distance measurement sensor (e.g. laser tracker). We just note that, the precision reached is relative to the measurement tool precision.

However, in future work, it will be useful to use an external measurement tool in order to compare our positioning method precision with other techniques. Indeed, the precision of the inverse kinematics of the robot cannot be defined with internal robot measurements. The previous statement is precisely the reason why we need to calibrate industrial robots. Hence, we will need to train the robot with external distance measurement sensors and to compare the precision with other methods using the same sensors.

\section{DISCUSSION}\label{discussion}

In previous work \cite{CDC16}, we showed that learning the cost function is more stable and converges faster than including distance in the state and approximate it in the first order. Here, we extend this work with second order improvement of the controller, which shows faster convergence properties under well chosen parameters. The high precision reached for this simple positioning task let us hope that such methods will be suitable for more complex industrial tasks. 

In many applications, it is also interesting to handle orientation of the end effector. Such modification is not an issue, one just needs to make several points on the end effector match several target points ($2$ or $3$ depending on the shape of the tool). This has been done with V-REP and converges just as well, even if taking more time. We chose not to present these results in this paper as they do not show any additional challenge and learning curves are less easy to interpret as distances are to be averaged between the three points.

In future work, we plan on trying to handle manipulation tasks with contact, which is a major challenge as the functions to approximate will not be smooth anymore near contact points.

\FloatBarrier
\bibliographystyle{IEEEtran}
\bibliography{IEEEabrv,maBiblio.bib}

\end{document}